\pdfoutput=1
\documentclass[11pt]{article}

\usepackage[]{ACL2023}
\usepackage{amsmath} % for text mode arrow
\usepackage{times}
\usepackage{latexsym}
\usepackage{pifont}
\usepackage{amssymb}
\usepackage[T1]{fontenc}
\usepackage[utf8]{inputenc}
\usepackage{bbding}
\usepackage{microtype}
\usepackage{graphicx}
\usepackage{booktabs}
\usepackage{tabularx}
\usepackage{multirow}
\usepackage{pgfplots}
\pgfplotsset{compat=1.18}
\usepackage{CJKutf8}
\usepackage{float}
\usepackage{subfigure}
\usepackage{hyperref} 
\newcommand{\improvement}[2]{%
  #1\textcolor{red}{\ensuremath{\,\uparrow\,}}\textcolor{red}{\footnotesize \textbf{#2}}
}

\title{PclGPT: A Large Language Model for Patronizing and Condescending Language Detection }

\author{Hongbo Wang\textsuperscript{\normalfont1}, Mingda Li\textsuperscript{\normalfont1}, Junyu Lu\textsuperscript{\normalfont1}, Hebin Xia\textsuperscript{\normalfont1}, Liang Yang\textsuperscript{\normalfont1}, Bo Xu\textsuperscript{\normalfont1}, Ruizhu Liu\textsuperscript{\normalfont2}, Hongfei Lin\textsuperscript{\normalfont1}\footnotemark[1]\\
         \textsuperscript{1}School of Computer Science and Technology, Dalian University of Technology, China\\
          \textsuperscript{2}Computer Science and Engineering College, Dalian Minzu University, China\\
         {\tt\{dutlaowang,22209217lmd,dutljy,2672054553\}@mail.dlut.edu.cn }\\
         {\tt \{liang,xubo,hflin\}@dlut.edu.cn}\\
         }

\begin{document}
\maketitle
\begin{abstract}
\textit{\textbf{Disclaimer}: Samples in this paper may be harmful and cause discomfort!}

Patronizing and condescending language (PCL) is a form of speech directed at vulnerable groups. As an essential branch of toxic language, this type of language exacerbates conflicts and confrontations among Internet communities and detrimentally impacts disadvantaged groups. Traditional pre-trained language models (PLMs) perform poorly in detecting PCL due to its implicit toxicity traits like hypocrisy and false sympathy. With the rise of large language models (LLMs), we can harness their rich emotional semantics to establish a paradigm for exploring implicit toxicity. In this paper, we introduce PclGPT \footnote{The data and code in this paper are available at \href{https://github.com/dut-laowang/emnlp24-PclGPT}{https://github.com/dut-laowang/emnlp24-PclGPT}.}, a comprehensive LLM benchmark designed specifically for PCL. We collect, annotate, and integrate the Pcl-PT/SFT dataset, and then develop a bilingual PclGPT-EN/CN model group through a comprehensive pre-training and supervised fine-tuning staircase process to facilitate implicit toxic detection. Group detection results and fine-grained detection from PclGPT and other models reveal significant variations in the degree of bias in PCL towards different vulnerable groups, necessitating increased societal attention to protect them.

\end{abstract}
\footnotetext[1]{* Corresponding author.}

%样例开始
\begin{CJK}{UTF8}{gbsn}
%数据集样例统计图
\begin{table*}[t]
\begin{center}
\begin{tabular}{ 
>{\centering\arraybackslash}m{7.0cm} 
>{\centering\arraybackslash}m{2cm} 
>{\centering\arraybackslash}m{1.5cm}
>{\centering\arraybackslash}m{1.5cm} 
>{\centering\arraybackslash}m{1.6cm} 
}
 \toprule[1.2pt]
 English Task
 & PCL Category
 & PLMs
 & GPT4.0
 & PclGPT-EN
 \\
 \midrule
  \textit {Since the elderly have been placed in a nursing home, they are undoubtedly left unattended most of the time. } & Unbalanced-Power-Relations & \XSolidBrush &\XSolidBrush &\Checkmark\\ 
\midrule
 Chinese Task
& PCL Category
 & PLMs
 & GPT4.0
 & PclGPT-CN
 \\
   \midrule
   战斗在火焰中激烈进行：茫然、饥饿的非洲难民在燃烧的大门中迷失方向。 & \multirow{4}{*}{Compassion} & \multirow{4}{*}{ \XSolidBrush} & \multirow{4}{*}{ \XSolidBrush}&\multirow{4}{*}{\Checkmark}\\  \textit {The fighting raged among the flames: Dazed, starving African refugees wandered lost through the burning portals.}     & & & & \\

\bottomrule[1.2pt]
\end{tabular}
\caption{PclGPT and other models' detection examples for ambiguous PCL. \ding{55} indicates incorrect prediction results, \ding{52} indicates correct prediction results.}
\label{tab:table1}
\end{center}
\end{table*}
\end{CJK}
%样例结束

\section{Introduction}

Patronizing and condescending language (PCL) specifically targets vulnerable groups. As an important but underexplored branch of toxic language, timely detection of PCL is crucial for protecting disadvantaged communities from further exclusion and inequality. Unlike traditional toxic languages such as hate speech \citep{cao2020hategan,caselli2020hatebert} and offensive language \citep{fortuna2020toxic,zampieri2019semeval,deng2022cold}, PCL expressions are more subtle and implicit (e.g., \textbf{\textit{" These poor children! It's truly admirable how they keep striving despite their humble beginnings."}}). This example is interesting because the original intention of PCL might have been to positively describe efforts to improve the lives of disadvantaged groups. However, it ultimately conveys subtle arrogance and discrimination, harming the individuals being sympathized with.
%细粒度分类开始
\begin{figure}[h]
\centering
\hspace*{-0.5cm}
\includegraphics[width=0.5\textwidth]{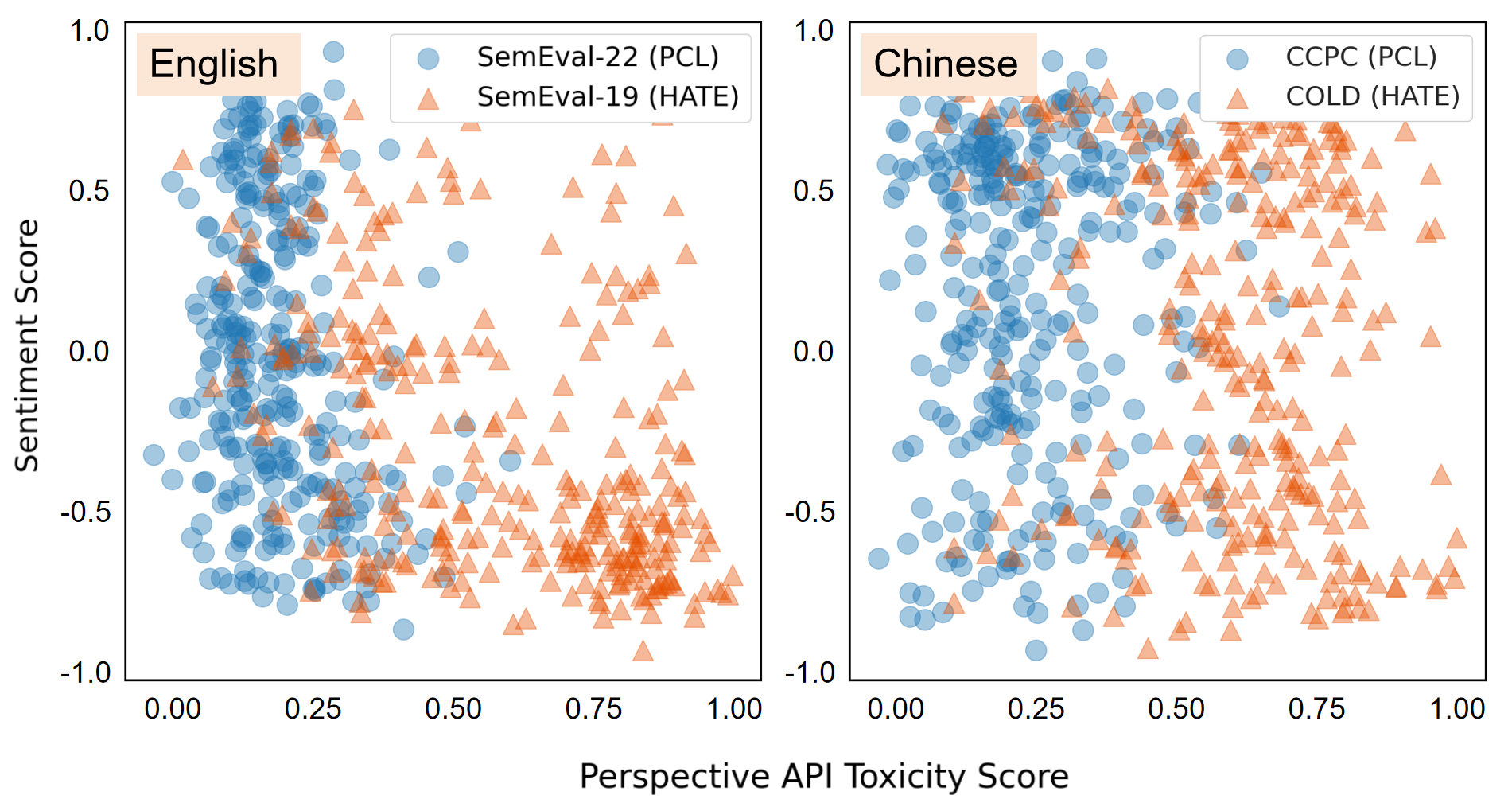}
\caption{Scatter plots for the scores using the Perspective API \citep{PerspectiveAPI} on the hate and PCL datasets. The left plot shows the English datasets SemEval-19 (HATE) and SemEval-22 (PCL), while the right plot shows the Chinese datasets COLD (HATE) and CCPC (PCL). The toxicity score ranges from 0 to 1, with increasing toxicity as discrete values.}
\label{fig:figure1}
\end{figure}
%细粒度分类结束

The subtle toxicity of PCL is further illustrated through toxicity scores. We compared the PCL and HATE datasets in both English and Chinese domains. As shown in Figure \ref{fig:figure1}, in both Chinese and English corpora, the toxicity scores of PCL are much lower than those of hate speech. This is due to the ambiguous toxic semantic features of PCL, which often lack explicit attacking vocabulary, leading to PLMs struggling to achieve optimal detection performance. The absence of high-quality data further constrains this field \citep{wang2023ccpc}. Large language models (LLMs) offer new opportunities with their extensive pre-trained knowledge and enhanced capability in revealing toxicity \citep{wen2023unveiling}. However, they still lack essential domain-specific knowledge for condescending language and effective guidance, leading to incomplete development for implicit toxic detection.

To address these challenges, we focus on three main questions: (1) How can we efficiently construct high-quality pre-training and supervised fine-tuning (SFT) datasets? (2) How can we design a new LLM benchmark that incorporates pre-training and SFT to enhance recognition of implicit toxicity? (3) Can we build a model group for other multilingual tasks like Chinese PCL detection to support vulnerable non-English-speaking communities?
%Our work has been publicly presented\footnote{\href{https://github.com/dut-laowang/PclGPT}{https://github.com/dut-laowang/PclGPT}}

To solve these issues, we introduce PclGPT, a comprehensive LLM benchmark for PCL detection, exploring the LLM's understanding of implicit toxicity. First, we collect community data from mainstream internet platforms (Reddit for English and Sina Weibo for Chinese) and process it to construct the Pcl-PT dataset for domain-adaptive pre-training. Next, we annotate, restructure, and filter high-quality data to construct the Pcl-SFT dataset, employing the instruction data paradigm to impose additional constraints on both input and output. Subsequently, we undertake the complete process of pre-training and SFT to construct our bilingual model, PclGPT-EN/CN. This model represents the first known LLM designed explicitly for PCL detection. Our results, shown in Table \ref{tab:table1}, illustrate the testing results on difficult-to-distinguish ambiguous examples. The model demonstrates superior performance compared to other PLMs and LLMs in both English and Chinese tasks.
Further group detection and fine-grained toxicity analysis reveal significant differences in the degree of bias in PCL towards various vulnerable groups. The ambiguity of bias also varies among different PCL subcategories. These findings necessitate increased societal attention to effectively protect vulnerable groups.

The main contributions of this paper are summarized as follows:

\begin{itemize}
    \item[$\bullet$] We construct the Pcl-PT/SFT datasets to enhance domain-specific knowledge for PCL. Pcl-PT is used for pre-training, covering over 1.4 million data entries from vulnerable communities. Pcl-SFT is used for fine-tuning, with high-quality bilingual instruction samples.
    \item[$\bullet$] We propose a pre-training and SFT framework to build our bilingual model, PclGPT-EN/CN. PclGPT is the first LLM designed to detect PCL and other implicit toxic languages, surpassing advanced PLMs and LLMs on four public datasets.
    \item[$\bullet$] Through group detection and fine-grained toxicity analysis, we demonstrate the differentiated nature of group biases in PCL, which means that biases against certain vulnerable groups require urgent attention, with PclGPT laying a foundation for managing these biases and protecting those groups.
\end{itemize}
\section{Related Work}

\textbf{Toxic Language.} Toxic language is perceived as an impolite, disrespectful, or irrational statement that may prompt someone to withdraw from a discussion \citep{dixon2018measuring}. Most existing research has concentrated on its largest subset — hate speech detection \citep{deng2022cold,caselli2020hatebert,mathew2021hatexplain,ocampo2023depth,bourgeade2023did,DBLP:conf/acl/LuXZMYL23,el-sayed-nasr-2024-aast-nlp}. However, hate speech typically focuses on direct attacks against specific groups based on religion, race, or ethnicity, while often neglecting other victims of toxicity, such as single-parent families, child laborers, and people with disabilities. Meanwhile, existing research equates toxic language with hate speech, focusing only on direct and explicit offenses and insults, while overlooking implicit forms of toxicity such as stereotypes and irony \citep{elsherief2021latent}. These gaps led to the emergence of PCL. 

\textbf{Implicit Toxic - PCL.} \citet{perez2020don} integrated categories of vulnerable groups and introduced PCL. This type of toxic language conveys a superior attitude or depicts vulnerable communities with pity or as needing help. Unlike traditional hate speech, PCL focuses on implicit toxicity aimed at marginalized and vulnerable groups. Such ambiguous implicit toxicity is less aggressive and has lower toxicity scores, which makes detection more challenging (Figure \ref{fig:figure1}). \citet{wong2014and} noted that PCL is often unconscious, driven by good intentions, and uses embellished language. \citet{xu2022xu} identified that such unjust treatment of vulnerable groups can exacerbate societal exclusion and inequality, causing users to leave communities or reduce online participation. While progress has been made in constructing PCL corpora \citep{wang2019talkdown,wang2023ccpc} and establishing specialized evaluation tracks, further research through improved deep learning networks continues \citep{perez2022semeval}, yet PCL detection still lacks comprehensive world knowledge. Their efficacy is significantly compromised by inadequate pre-training and the implicit nature of toxicity within PCL.

\textbf{LLM for Toxicity Detection.} In recent years, decoder-only LLMs, such as ChatGPT \citep{openai2022chatgpt}, GPT-4 \citep{openai2023gpt}, and LLaMA \citep{touvron2023llama}, have revolutionized text generation. LLMs have increasingly been applied in toxic language detection and prevention. \citet{shaikh2022second} demonstrated that zero-shot CoT significantly increases LLMs' toxic output. \citet{wen2023unveiling} proved that SFT and reinforcement learning further induce toxic outputs. \citet{zhu2023can,huang2023chatgpt} used ChatGPT to map answers to binary labels through prompt engineering for hate detection. \citet{roy2023probing} enhanced hate speech classification accuracy by including additional victim information. However, no systematic LLM engineering is currently used to detect PCL or other discriminatory texts. Additionally, LLMs' fine-grained discrimination of implicit toxicity remains vague. To address these gaps, we introduce PclGPT, a dedicated LLM benchmark for PCL detection, which leverages pre-training and SFT to surpass existing models on four public datasets.
%主图开始
\begin{figure*}[t]
\centering
\includegraphics[width=0.9\textwidth]{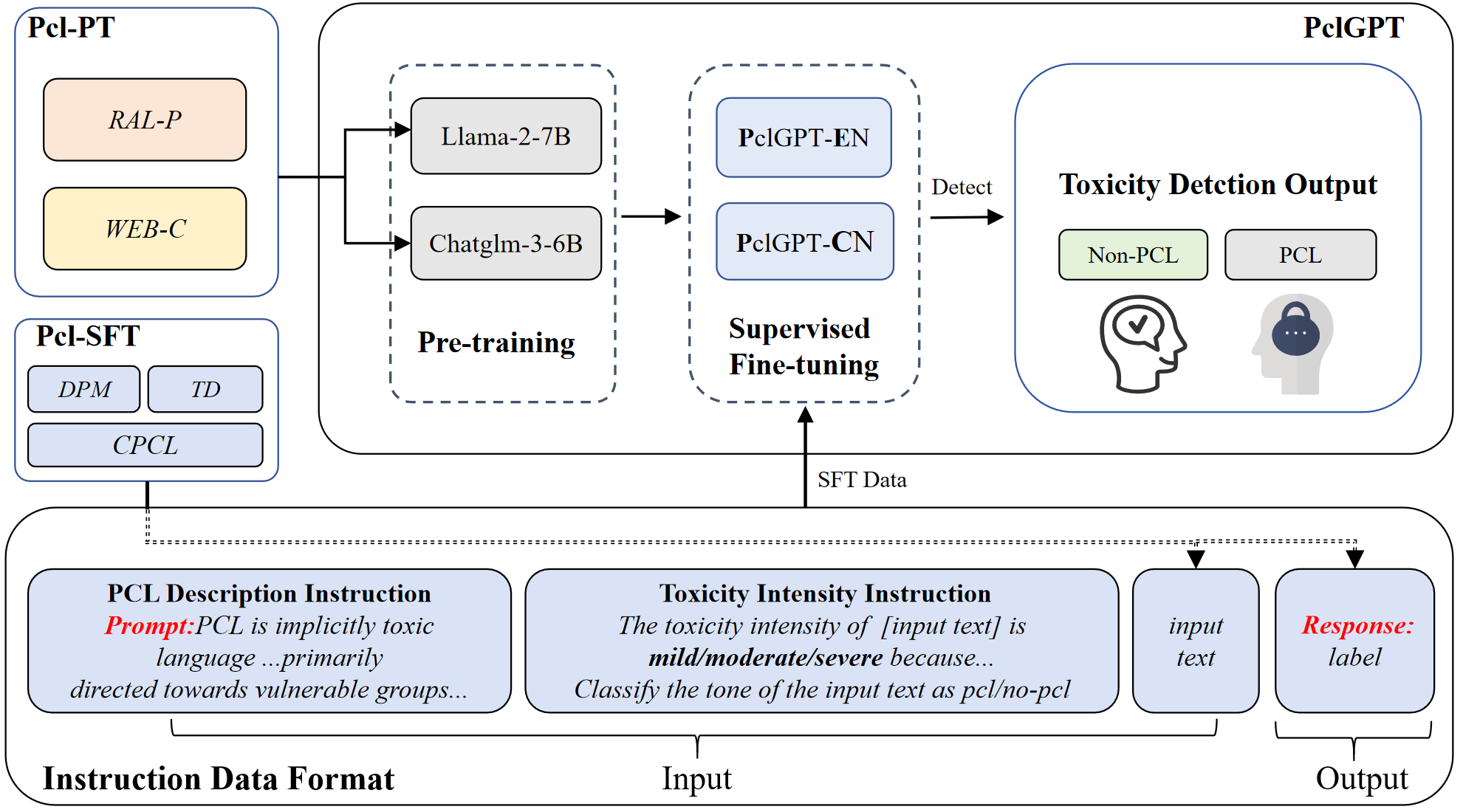}
\caption{ An illustration of the overall PclGPT. We establish Pcl-PT/SFT datasets and build a bilingual model group through pre-training and SFT. Instruction Data Format demonstrates the data construction format for SFT.}
\label{fig:figure2}
\end{figure*}
%主图结束

\section{PclGPT}

The overall approach is illustrated in Figure \ref{fig:figure2}. Our PclGPT model group consists of two sub-models: PclGPT-EN and PclGPT-CN, using LLaMA-2-7B and ChatGLM-3-6B \citep{du2022glm} as their base architectures, respectively. LLaMA, one of the foremost English open-source LLMs today, has been pre-trained on over 20 trillion tokens. ChatGLM, among the most advanced Chinese LLMs, is built upon the Generalized Linear Model (GLM) architecture and has been extensively optimized for Chinese question-answering and dialogue tasks, exhibiting outstanding performance in the Chinese domain. LLaMA-2-7B has a context length of up to 4096 tokens and ChatGLM-3-6B with 8192 tokens, ensuring a thorough understanding of the context. Detailed descriptions of the pre-training and fine-tuning stages will be provided in the subsequent sections.

\subsection{Pre-training}
\label{Sec:sec3.1}
To facilitate the pre-training process, we introduced the Pcl-PT dataset, comprising the RAL-P and WEB-C datasets. Specifically, we employed separate corpora in English and Chinese to pre-train our PclGPT-EN/CN model group. Our pre-training followed a standard paradigm, where the model predicted the next token based on existing input history. For both PclGPT-EN and PclGPT-CN, we utilized the same vocabulary as the base models and employed AdamW as the optimizer. The initial learning rate was set to $2 \times 10^{-4}$ with a weight decay of 0.1. We also employed efficient training strategies, including mixed precision training with bf16 \citep{micikevicius2017mixed}. The specific parameters are detailed in Appendix \ref{app:A}. Below, we provide detailed insights into the datasets. The design of our dataset follows the hierarchical format of \citet{tian2023chimedgpt}, with more details shown in Table \ref{tab:table2}.

\begin{itemize}

\item[$\bullet$] \textbf{RAL-P} is derived from the RAL-E dataset. The RAL-E dataset \citep{caselli2020hatebert} includes offensive, abusive, and hateful content from the Reddit community, comprising 43M tokens collected from December 2005 to March 2017. However, RAL-E predominantly features explicit hate speech, which hinders the accurate identification of PCL, as the toxicity of PCL is often not directly correlated with explicit intensity, positive samples may also convey biased intentions. Therefore, based on the criteria established by \citet{perez2020don}, we used LLM to generate a dictionary of over 500 condescending English terms, which was manually calibrated by three proofreaders who collaboratively filtered out terms unrelated to PCL, ultimately retaining 379 relevant terms. We used this dictionary to match RAL-E with data more closely related to PCL, while retaining 30\% of non-PCL entries to ensure balanced pre-training data. RAL-P ultimately retained 1091945 data entries. Detailed processes are presented in Appendix \ref{app:b}.

\item[$\bullet$] \textbf{WEB-C}. The scarcity of data in the Chinese domain constrains the task of PCL detection. To address this, we designed a framework to systematically gather bullying, violent, and discriminatory content from marginalized communities on Sina Weibo, a mainstream Chinese media platform. We initially limited the search scope to eight major disadvantaged group categories based on PCL criteria \cite{wang2023ccpc}, and expanded the keyword list accordingly. We then crawled Weibo posts from July 2022 to January 2024 using these keywords and performed data filtering and user-sensitive information replacement. Ultimately, we collected 315074 instances. The detailed keyword list and data collection process are presented in Appendix \ref{app:b}.
\end{itemize}
\begin{table*}[t]
\centering
\setlength{\tabcolsep}{5pt} 
\begin{tabular}{l|l|l|l|l}
\toprule[1.2 pt]
\textbf{Stage}            & \textbf{Dataset}    & \textbf{Language} & \textbf{Method}     & \textbf{\#Instances}  \\ \midrule[1.2 pt]
\multirow{2}{*}{Pcl-PT} & RAL-P                 & EN       & \textbf{Self-built} & 1091945                     \\ 
                              & WEB-C                   & CN       & \textbf{Self-built} &  315074                \\ \midrule
\multirow{3}{*}{Pcl-SFT}          & Don't Patronize Me (DPM) & EN       & Public     & 10469                         \\             & TalkDown (TD)            & EN       & Public     & 74865            
                      \\        & CPCL                    & CN       & \textbf{Self-built} & 18253                  \\ \midrule
Test                          & DPM/TD/CPCL/CCPC             & EN,CN    & Public     &   N/A                 \\ \bottomrule[1.2 pt]

\end{tabular}
\caption{Statistics of the datasets used in training PclGPT under different stages.  Pcl-PT is used in the pre-training stage, and Pcl-SFT is used in the SFT stage. "Method" means we construct our own dataset / modify a public corpus. "Instances" represents the number of sentences or texts.}
\label{tab:table2}
\end{table*}
\subsection{Instruction Data Format}
\label{Sec:sec3.2}
\begin{figure}[h]
\centering
\includegraphics[width=0.45\textwidth]{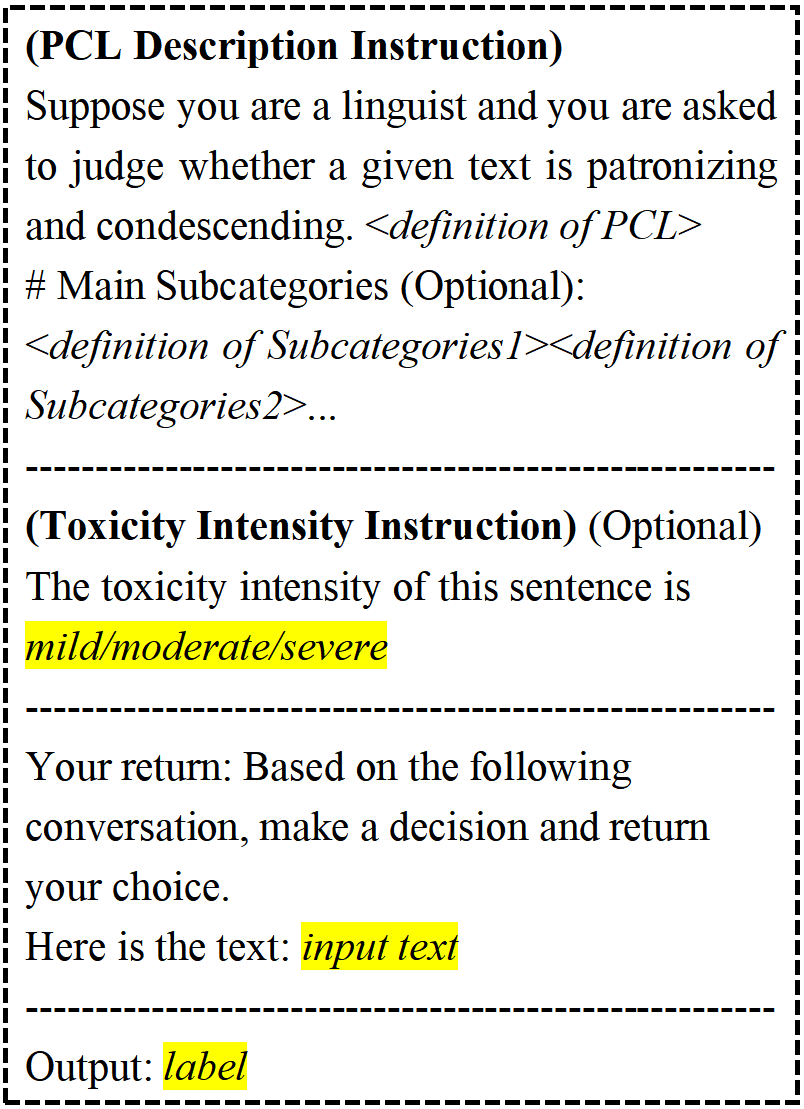}
\caption{A template for SFT instructions, including definitions of PCL and its subcategories, as well as toxicity intensity.}
\label{fig:figure3}
\end{figure}

Recent studies have underscored the critical role of SFT in shaping the cognitive capabilities of LLMs, with properly formatted instruction data aiding in fully leveraging the knowledge potential of LLMs \citep{taori2023stanford,chiang2023vicuna,ouyang2022training}. It has been pointed out that incorporating fine-grained toxicity intensity can further enhance the robustness of PCL recognition \citep{wang2023ccpc}. The instruction templates we constructed include both \textit{\textbf{PCL Description Instruction}} and \textit{\textbf{Toxicity Intensity Instruction}}, designed to more accurately capture the implicit semantic characteristics of PCL, as shown in Figure \ref{fig:figure3}.

\textbf{\ PCL Description Instruction.} Since PCL is a subjective toxic category, first, we need a complete description of PCL to guide the model to respond in a standardized format. The description includes the definition and subcategories. This part of the content is fixed and descriptive.
%prompt开始
%prompt结束

\textbf{\ Toxicity Intensity Instruction (optional).} Next, we focus on the potential influence of the intensity of toxicity on implicit emotions. We used the open-source Perspective API to score the text for toxicity and based on these scores, we integrated toxicity intensity labels into the original data, categorizing them as mild, moderate, and severe.

\subsection{Supervised Fine-tuning}
Following the instruction format outlined in Section \ref{Sec:sec3.2}, we constructed the Pcl-SFT dataset for the SFT process, comprising English datasets: Don’t Patronize Me! \citep{perez2020don} and TalkDown \citep{wang2019talkdown}, as well as the Chinese dataset CPCL. We adhered to the same bilingual training rules described in \ref{Sec:sec3.1} to ensure the multilingual detection capability of PclGPT. In the following sections, we present detailed information regarding the Pcl-SFT dataset. More details are shown in Table \ref{tab:table2}.
 \begin{itemize}
    \item[$\bullet$] \textbf{Don’t Patronize Me! (DPM)}  contains 10,469 English paragraphs about potentially vulnerable groups, extracted from the News on the Web (NoW). The dataset was annotated hierarchically with numerical labels ranging from 0 to 4, indicating the toxic intensity of PCL. In SFT, we only utilized information from community texts and their corresponding labels.
     %各类别统计图
\begin{table*}[htbp]
\begin{center}
 \vspace{-0.3cm}
\begin{tabular}{ 
>{\centering\arraybackslash}m{0.8cm} 
>{\centering\arraybackslash}m{1.3cm} 
>{\centering\arraybackslash}m{1.3cm} 
>{\centering\arraybackslash}m{1.3cm}
>{\centering\arraybackslash}m{1.3cm} 
>{\centering\arraybackslash}m{1.6cm} 
>{\centering\arraybackslash}m{1.3cm} 
>{\centering\arraybackslash}m{1.3cm}
>{\centering\arraybackslash}m{1.3cm}
}
 \toprule[1.2pt]
 & Disabled & Women& Elderly &Children&Single-parents  &Ordinary.&Disadv. groups
 &Total\\ 
 \midrule
 zhihu & 1208 & 1147 &1131& 1619 &1113&1093&1959&9270\\ 
 zhihu$_p$   & 338  & 248 &294  & 374 & 264& 263& 354& 2135\\ 
 prop.(\%)  &28.0  &21.6 &26.0 &23.1&23.7&24.1&18.1&23.0\\ 
 \midrule
 weibo & 1102&974&1247 &1588& 1077&944&2051&8983\\ 
 weibo$_p$ &310&241&267 &592& 389&123&533&2455\\ 
 prop.(\%)  &28.1 &24.7 &21.4& 37.3 &36.1&13.0&26.0&27.3\\ 
 \midrule
 Total  &2310 &2121 &2378&3207&2190&2037&4010&18253\\ 
\bottomrule[1.2pt]
\end{tabular}
\caption{Statistical Results of CPCL from different Platforms. Platform$_{\mbox{\scriptsize p}}$ represents samples marked as PCL, whereas prop.(\%) represents a percentage.}
\label{tab:tableS}
\end{center}
\end{table*}
 \vspace{-0.3cm}
    \item[$\bullet$] \textbf{TalkDown (TD)} is a Reddit community dataset containing 74K English comment/reply pairs. The collected information comes from disadvantaged groups from 2006 to 2018. Each pair is marked as one of three categories: PCL, non-PCL, and unsure. In SFT, we concatenated comment/reply pairs and manually filtered a subset for training. An offensive language dictionary was applied to remove aggressive pairs, aligning with PCL's less offensive nature. To ensure model fairness, data exceeding the input limitations for long texts were discarded.
%各类别统计图 
    \item[$\bullet$] \textbf{CPCL} is a Chinese dataset we manually collected and annotated from Chinese social media platforms. We conducted hierarchical structured annotations on the data according to the toxicity definition of PCL \cite{perez2020don,wang2023ccpc}. The annotations include toxicity existence, fine-grained PCL categories, and considerations for vulnerable groups. The corpus now has more than 18K two-level structured annotations. Detailed statistics of the CPCL dataset, categorized by media platform and targeted towards vulnerable communities, are shown in Table \ref{tab:tableS}. For toxicity categories, we used Wang's standard \cite{wang2023ccpc} to categorize Chinese PCL statements into the following subcategories: \textit{“Unbalanced Power Relations”}, \textit{“Spectator”}, \textit{“Prejudice”}, \textit{“Appeal”}, and \textit{“Elicit Compassion”}. The annotation process involved specialized training, with two annotators for the initial annotation and one annotator for proofreading, to minimize subjective errors in marginal cases. Additionally, we performed a subjective consistency review on the annotation results to ensure the reliability of our annotated data. Appendix \ref{app:c} describes a more detailed annotation process.

\end{itemize}

We transformed the union of the original datasets into the SFT data format, combining PCL descriptions with toxicity intensity as described in Section \ref{Sec:sec3.2}. We connected pairs of Enhancement-Response to form long input texts, maximizing the sequence length of LLMs. During training, we used sequence-to-sequence loss exclusively and map the final generated output to binary label pairs. We performed SFT on 8 RTX 4090 GPUs, conducting 5 epochs of full-parameter tuning with the AdamW optimizer at a learning rate of 2e-5. The specific parameters are detailed in Appendix \ref{app:A}.

\subsection{Bias Detection for PCL}
Inspired by \citet{zhang2023instruct}, we further investigated the effectiveness and fairness of our PclGPT model through group detection and fine-grained classification tasks.

\textbf{Group Detection.} Group detection helps us address bias issues in the model against different demographics. We conducted experiments using the DPM dataset, which balances coverage across various minority groups. We compared fine-tuned BERT series models with PclGPT-EN in these experiments.

\textbf{Fine-Grained Analysis.} Fine-grained analysis of toxicity categories is crucial for understanding implicit toxic sentiments \citep{tang2019aspect}. Our Chinese CPCL dataset divides PCL into five subcategories. We split the CPCL dataset into five subsets based on these categories to test the sensitivity of PclGPT-CN to different toxicity types. We compared PclGPT-CN with Chinese-BERT \citep{sun2021chinesebert} and ChatGLM in these experiments.

%主实验表开始
% \usepackage{multirow}
\begin{table*}[t]
\setlength{\tabcolsep}{4.5pt} 
\begin{tabular}{ll|llllll|lll|c} % 左边三列的宽度增加
\toprule[1.2 pt]
\multicolumn{2}{l|}{}                                    & \multicolumn{3}{c|}{\textbf{DPM}}        & \multicolumn{3}{c|}{\textbf{TD}} & \multicolumn{3}{c|}{\textbf{CPCL }} & \textbf{CCPC} \\ \cmidrule{3-12} 
\textbf{LM }                                       &\textbf{Model}        & P & R & \multicolumn{1}{l|}{F1} & P      & R     & F1     & P       & R       & F1      & \text{F1} \\ \midrule[1.1 pt]
\multirow{4}{*}{PLMs} & RoBERTa      & 76.3 &78.7 &\multicolumn{1}{c|}{77.4 }  &  88.4 &86.7 & 86.5       & 61.2  & 61.3  & 61.3        & 55.4 \\
\multicolumn{1}{c}{}                      & RoBERTa-L    & \underline{80.2}  & 74.9  & \multicolumn{1}{c|}{77.2}   & 88.1  & 86.0 & 85.9  & 62.5  & 61.6   & 62.0        & 55.3 \\
\multicolumn{1}{c}{}                      & Chinese-BERT  & 71.2  & 63.5  & \multicolumn{1}{c|}{66.2}   &  76.7 &74.7&74.2  &  66.6  & \underline{71.0}  &67.3 &    57.1        \\
\multicolumn{1}{c}{}                      & M-BERT  & 69.2  &76.0   & \multicolumn{1}{c|}{71.8}   &    87.6 &87.4 &87.4 &65.8  &67.8    &66.6         & 56.0\\ \cmidrule{1-12} 
\multirow{5}{*}{Base-LLMs}   & ChatGPT      &50.8  &52.3   & \multicolumn{1}{c|}{46.9}   & 59.2  &58.1  &56.7  & 53.1&54.2 &53.6        & 53.3 \\
         & GPT-4.0      & 51.5  & 57.5  & \multicolumn{1}{c|}{54.3}   & 60.8 & 60.3   & 60.5  &  55.4       &  56.3     & 55.7    & 56.3 \\
        & Claude-3      & 52.3  & 52.5  & \multicolumn{1}{c|}{52.3}   &61.6 &  64.1  & 63.2 &57.2 &57.7  &57.3        & \underline{57.6} \\ \cmidrule{2-12} 
     & LLaMA-2-7B   & 50.9  &52.6   & \multicolumn{1}{c|}{51.4}   & 49.9 &  49.9  &  49.7 &  45.2  &47.5   &46.3        & 42.5 \\
    & ChatGLM-3-6B & \textcolor{gray}{N/A}  &\textcolor{gray}{N/A}   & \multicolumn{1}{c|}{\textcolor{gray}{N/A}}  &\textcolor{gray}{N/A}  &\textcolor{gray}{N/A}  &\textcolor{gray}{N/A} &    51.9 &50.2  & 51.0       & 49.1\\ \cmidrule{1-12} 
        \multirow{4}{*}{LLMs(Ours)}& \textbf{PclGPT-EN}    & \textbf{80.4}  & \textbf{81.8}  & \multicolumn{1}{c|}{\textbf{81.1}}   & \textbf{89.9} &   \textbf{89.0}    &   \textbf{88.9}     & \textcolor{gray}{N/A}&\textcolor{gray}{N/A} &\textcolor{gray}{N/A} & \textcolor{gray}{N/A} \\ 
         & \textit{- TII}   &  79.5 & \underline{80.3} & \multicolumn{1}{c|}{\underline{79.9}}   &  \underline{88.5}  &   \underline{88.0}  & \underline{88.2} &   \textcolor{gray}{N/A}  & \textcolor{gray}{N/A}   & \textcolor{gray}{N/A} & \textcolor{gray}{N/A} \\ 
         & \textbf{PclGPT-CN}   &  \textcolor{gray}{N/A} &  \textcolor{gray}{N/A} & \multicolumn{1}{c|}{\textcolor{gray}{N/A}}   & \textcolor{gray}{N/A}  &  \textcolor{gray}{N/A}  & \textcolor{gray}{N/A} &   \textbf{69.1}  & \textbf{72.0}   & \textbf{70.2} & \textbf{60.2} \\ 
         &  \textit{- TII}    &  \textcolor{gray}{N/A} &  \textcolor{gray}{N/A} & \multicolumn{1}{c|}{\textcolor{gray}{N/A}}   & \textcolor{gray}{N/A}  &  \textcolor{gray}{N/A}  & \textcolor{gray}{N/A} &   \underline{68.1}  &  71.0   & \underline{69.5} & 57.2 \\ \bottomrule[1.2 pt]
\end{tabular}
\caption{The results indicate the macro-average precision (P), recall (R), and F1-score, calculated by weighting the F1 of positive and negative samples. Optimal and suboptimal scores are denoted in \textbf{bold} and \underline{underlined}, respectively. For optimal performance, we used the model test data for each language, with "N/A" for non-applicable segments. CPCL is our new Chinese dataset, while CCPC \citep{wang2023ccpc} serves as a comparative experiment to validate the generalization ability of CPCL. \textit{- TII} is the result of removing the Toxicity Intensity Instruction template.}
\label{tab:table3}
\end{table*}
%主实验表结束

\section{Result and Analysis}
\subsection{Baselines and Settings}
To validate the performance of PclGPT, we extensively tested various PLMs and LLMs with our PclGPT model group on four public datasets (two in English and two in Chinese). To ensure our model demonstrates the best performance on bilingual PCL detection, we used PclGPT-EN to detect the English datasets and PclGPT-CN for Chinese.

\textbf{PLMs.} Pre-trained language models have consistently been the most important types of models in traditional toxicity detection tasks. We employed BERT and its relevant variants within the PLM category, such as RoBERTa \citep{liu2019roberta}, Chinese-BERT, and Multilingual-BERT (M-BERT) \citep{pires-etal-2019-multilingual}. To ensure the optimal performance of PLMs on the test set, we used the standard training and fine-tuning workflow. The training portions of three public datasets were used for training the PLMs (For CCPC, we continued using the training set of CPCL). Additionally, both PLMs and LLMs were evaluated using the same test set to ensure comparability. Detailed parameters are shown in Appendix \ref{app:A}, providing comprehensive insights into our experimental setup.

\textbf{Base-LLMs.} The use of LLMs is divided into two parts. For advanced but non-open-source LLMs, such as ChatGPT and Claude-3 \citep{claude3}, we accessed them via API calls. Meanwhile, we used the original versions of LLaMA-2-7B and ChatGLM-3-6B without any parameter fine-tuning as part of the PclGPT ablation study to evaluate the performance improvements. To ensure experimental consistency, we used the same instruction format for other LLMs as used for PclGPT. Given that PCL represents implicit toxicity, and the performance of base LLMs with few-shot setups remains limited, we employed zero-shot testing for a clearer comparison.

For the results of both PLMs and LLMs, we evaluated the models using macro-average precision (P), recall (R), and F1-score (F1).

\subsection{Overall Performance}
Table \ref{tab:table3} compares the performance of PclGPT with PLMs and other LLMs on four test sets.
\begin{itemize}
 \item[$\bullet$] PLM still holds significant importance in the field of toxicity detection, but the disadvantages are apparent. From the perspective of subjective ambiguity, PLM performs well on the Talkdown (English) dataset, which has a uniform data distribution and clear definitions. However, it performs poorly on the DPM (English) and CPCL (Chinese) datasets, where the definition of condescension is more ambiguous.
%细粒度分类开始
\begin{figure*}[htbp]
\centering
\includegraphics[width=0.9\textwidth]{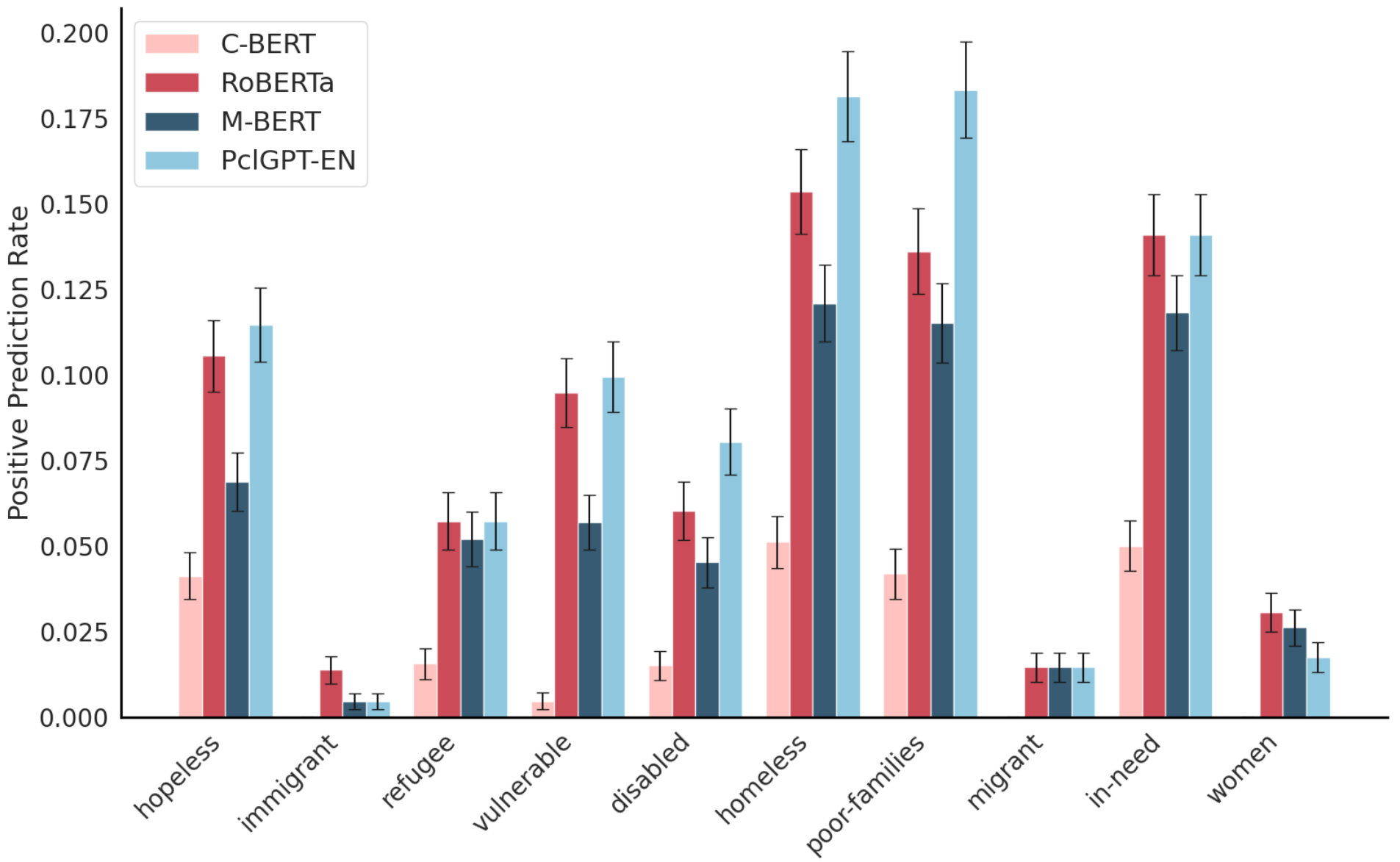}
\caption{Group detection for different models. The test group consists of 10 different disadvantaged communities.}
\label{fig:figure4}
\end{figure*}
%细粒度分类结束
 \item[$\bullet$] PclGPT has achieved outstanding results in both English and Chinese domains, with particularly noticeable improvements in detecting ambiguous data. Specifically, PclGPT improved by 3.7\% on the DPM dataset compared to the best RoBERTa model, and by 2.9\% on the CPCL dataset compared to the best Chinese-BERT model.

 \item[$\bullet$] Base-LLMs, without parameter adjustments, have not realized their potential in subjective toxicity detection. Due to insufficient emphasis on toxic texts, unadjusted LLMs show low performance in detecting implicit toxic texts like PCL. Compared to PLMs, LLMs' average performance drops by about 20.49\% in precision, 18.87\% in recall, and 19.66\% in F1 score. This drop is intriguing as PCL samples often contain positive expressions and goodwill, interfering with LLMs' pre-trained features. PclGPT effectively guides LLMs in understanding PCL toxicity definitions and subcategories, providing essential guidelines for future LLM safety regulations.
\end{itemize}

\subsection{Result for PCL Group Detection}

In Figure \ref{fig:figure4}, we compared the performance of PclGPT-EN and other models in detecting PCL across different vulnerable groups. The test set had an even distribution of various vulnerable groups and positive samples. However, the models showed a clear preference for identifying poor-families and homeless individuals, indicating that these groups exhibit more identifiable semantic features. Expressions of sympathy and pity towards these groups are more likely to be perceived as condescending. PclGPT further enhanced the detection capability for these groups. In contrast, ambiguous discriminatory attitudes towards migrants and immigrants remain challenging to identify, suggesting that additional measures are necessary to protect these groups.
%细粒度测试图
\begin{table}[htbp]  % 使用 [H] 强制表格放置在当前位置
\centering
\begin{tabular}{p{0.25\linewidth} p{0.15\linewidth} p{0.18\linewidth} p{0.2\linewidth}}  % 调整每列的宽度，确保表格适合单栏
\hline
   \textbf{Category} & \textbf{Chat-GLM} & \textbf{Chinese-BERT} & \textbf{PclGPT-CN} \\
\hline
        
        Unb. & 52.1 & 66.5  & \textbf{69.4}\improvement{}{2.9}  \\
       Spectators & 44.3  & 71.3 & \textbf{72.1}\improvement{}{0.8} \\
       Prejudice & 49.7&  64.3& \textbf{67.5}\improvement{}{3.2} \\
        Appeal & 24.5 & 59.0& \textbf{65.0}\improvement{}{6.0} \\
       Compassion &44.2  & 52.3 &\textbf{57.4}\improvement{}{5.1} \\

\hline
\end{tabular}
\caption{Experimental results for fine-grained PCL Detection. We evaluated our model using the macro-average F1-score as the metric. }
\label{tab:table4}  
\end{table}
\subsection{Result for Fine-grained PCL Detection}

 Table \ref{tab:table4} presents the results of our fine-grained PCL testing. Our experiment indicated that models still exhibit varying degrees of bias in detecting different subcategories of PCL. In the "Appeal" and "Compassion" subcategories, subjective and ambiguous expressions effectively evade the recognizer's correct functioning. Notably, our PclGPT-CN showed improved performance across all subcategories, with the most significant improvement in the ambiguous "Appeal" subcategory.

\section{Conclusion }

In this paper, we introduce PclGPT, a large language model group designed to detect PCL targeting vulnerable groups. As a subset of the toxic language, PCL harms vulnerable groups through discriminatory language. Traditional PLMs struggle with PCL detection due to its implicit harmful features. PclGPT significantly improves detection performance by leveraging the emotional semantic capabilities of LLMs. We collect, annotate, and merge the Pcl-PT/SFT dataset, and establish the PclGPT-EN/CN through comprehensive pre-training and SFT process to detect PCL in both Chinese and English communities. PclGPT outperforms existing models on four public datasets, demonstrating its strong capability in handling implicit harmful language. Additionally, group detection and fine-grained toxicity analysis reveal significant bias differences against various vulnerable groups, highlighting the urgent need for societal protection. PclGPT's development enhances PCL recognition and provides new directions and tools for future implicit toxic language research.

\section{Limitation }

PCL is a subclass of microaggressions within toxic language. Due to the limited research in this area, further linguistic foundation is needed to refine the standardized definition of this type of speech. Our current study lacks an examination of “false positive” cases, such as insincere benevolence and superficial compliments directed at marginalized communities. Moreover, because of its implicit toxic nature, research on PCL can substantially contribute to the study of other forms of implicit toxicity or aggression, such as implicit hate speech, sarcasm, and stereotypes, guiding our future research. Considering the potential for toxic optimization and value-based controversies when using reinforcement learning from human feedback (RLHF) in training models, we did not apply RLHF in this paper. For further details, please refer to Appendix \ref{app:g}.

\bibliography{anthology,ref}
\bibliographystyle{acl_natbib}

\appendix
\section{Parameter Settings}

\label{app:A}

In this section, we provided a detailed description of the experimental parameter settings. This included the parameters for pre-training and SFT of the PclGPT.
%参数图
\begin{table*}[t]
  \centering
  \begin{tabularx}{1.0\linewidth}{p{0.25\linewidth}p{0.2\linewidth}p{0.25\linewidth}p{0.3\linewidth}}
\hline
 \textbf{Parameter\_for\_PT}&\textbf{Value} &\textbf{Parameter\_for\_SFT}&\textbf{Value}\\
 \hline
 Lr&2e-4&Lr&2e-5\\
 Batchsize &32& Batchsize&16\\
Training Epochs&5&Training Epochs&5\\
 Max Source Len&512&Block Size&1024\\
 Max Target Len&512&-&-\\
GPUs & RTX 4090*8 (24G)&GPUs & RTX 4090*8 (24G)\\
-&- &GPUs\_inference &A100\_PCIE*2 (40G) \\
\hline 
\end{tabularx}
\caption{Detailed configuration parameters for the pre-training and SFT phases of PclGPT. The inference phase uses the same GPU configuration as the PLM test.}
\label{tab:table6}
\end{table*}
\subsection{PLM Settings}
To compare our PclGPT, we fine-tuned our PLMs using the same size training and test sets as those used for PclGPT. Specifically, we conducted fine-tuning experiments for 5 epochs on 2 A100 GPUs and used the best epoch model weights for test set predictions. We tested RoBERTa, Chinese-BERT, and M-BERT models on four datasets. The specific parameters are as shown in Table \ref{tab:table5}.
\begin{table}[H]
\begin{tabularx}{1.0\linewidth}{p{0.5\linewidth}p{0.5\linewidth}}
\hline
\textbf{Parameter\_for\_PLM} & \textbf{Value}             \\
\hline
Lr                  & 1e-2              \\
Max\_len            & 1024              \\
Batchsize     & 16                 \\
Training Epochs     & 5                 \\
warmip\_steps       & 500               \\
GPUs                & A100\_PCIE*2 (40G) \\
\hline
\end{tabularx}
\caption{Detailed parameter settings for the fine-tuning and testing phases of PLMs.}
\label{tab:table5}
\end{table}

\subsection{PclGPT Settings}
For PclGPT, due to the scale effect of the pre-training corpus, we set a higher learning rate and batch size than SFT. Both the pre-training and SFT were conducted on 8 RTX 4090 GPUs. We accomplished this procedure and guaranteed the consistency of the pertinent training parameters in both Chinese and English models. During the inference phase, to control for a single variable, we used the same configuration of 2 A100 GPUs as in the PLM fine-tuning, as shown in Table \ref{tab:table6}. This inference setup is also applicable to the zero-shot inference process for non-API Base-LLMs, like LLaMA-2-7B and ChatGLM-3-6B.

\section{Detailed Construction of the Pcl-PT Dataset}
\label{app:b}
\begin{figure}[h]
\centering
\includegraphics[width=0.45\textwidth]{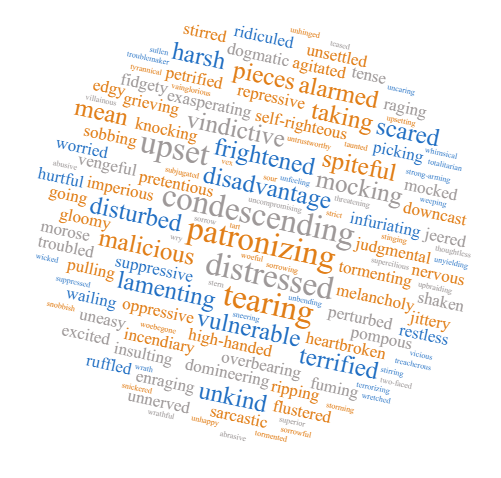}
\caption{Word cloud statistics of the condescending dictionary.}
\label{fig:figure5}
\end{figure}

\textbf{RAL-P.} In the process of transforming RAL-E, we used LLM to construct a PCL dictionary. Specifically, we had the LLM generate over 500 words that best reflect patronizing semantics based on confidence levels, which were then manually verified. Part of the word cloud information sorted by confidence levels is shown in Figure \ref{fig:figure5}. For sentences in RAL-E that did not contain any dictionary information, we retained only 30\% as non-patronizing corpus, while all sentences containing any dictionary information were retained. The original text corpus consisted of 1,476,472 sentences, and the filtered corpus contained 1,091,945 sentences, which were used as RAL-P pre-training data.
%词云开始
%词云结束

\begin{CJK}{UTF8}{gbsn}
\textbf{WEB-C.} We uniformly collected data from eight common vulnerable groups on the Weibo platform as our WEB-C Chinese pre-training corpus. The annotation team added 20 of the most commonly used search terms for each group, resulting in the final search list. Detailed information on community categories can be found in Table \ref{tab:tableB}. For filtering, we removed duplicate and irrelevant samples (including common fixed tags on Weibo such as "\# 话题内容" and "\# 评论日期"), and we replaced user information with \#USER to comply with the community privacy agreement. We retained the emojis in the samples and converted them to the corresponding Chinese text specified by the platform to preserve as much of the emotional semantic information conveyed by the emojis as possible.
\end{CJK}

%WEB-C
\begin{table}[h]
\centering
\begin{tabularx}{0.9\linewidth}{p{0.5\linewidth}p{0.4\linewidth}}
\hline
\textbf{Community} & \textit{Total}  \\
\hline
        
        \# Disabled & 38981   \\
        \# Women  & 40256 \\
        \# Elderly  & 39385 \\
        \# Children & 38475   \\
        \# Single-parent  & 40689 \\
        \# Ordinary People  & 37589 \\
        \# Disadvantaged  & 40324 \\
        \# Others  & 39375 \\

\hline
\end{tabularx}
\caption{The final collection status of different PCL communities. }
\label{tab:tableB}
\end{table}
%WEB-C

\section{Detailed Construction of the Pcl-SFT Dataset}
\label{app:c}
\textbf{CPCL.} We adopted the same method as WEB-C described in Appendix \ref{app:b} for data selection and filtering, and manually annotated the high-quality texts. This section provides a detailed description of the annotation and statistics of our constructed CPCL dataset. Due to the subjective nature of PCL speech, we abandoned the automatic annotation method by LLM and continued to use manual annotation. We recruited four annotators with diverse gender, age, and educational backgrounds (two primary annotators and two proofreaders) (50\% female, 50\% male; age 25±5 years; two master's degree holders, two PhD holders). We adopted the standard proposed by \citet{wang2023ccpc} and conducted detailed training on test samples before annotation to ensure that annotators understood the subtle toxicity differences of PCL. The annotation was uniformly conducted using the annotation template as shown in Figure \ref{fig:figure8}. To ensure annotation consistency, we calculated the Kappa inter-annotator agreement (IAA) for binary and multi-class annotations. The IAA results are shown in Table \ref{tab:table8}. If we ignore all annotations marked as low toxicity intensity by at least one annotator, the IAA improves. This indicates that PCL with weak toxicity intensity has higher ambiguity. 
%IAA
\begin{table}[htbp]
\centering
\begin{tabularx}{0.9\linewidth}{p{0.5\linewidth}p{0.4\linewidth}}
\toprule[1.2pt]
\textbf{Binary-classification} & \textit{Kappa IAA}  \\
\midrule
        
         All labels & 0.62   \\
         Remove Weak level   & 0.67 \\
\midrule
\textbf{Multi-classification} & \textit{Kappa IAA}  \\
\midrule

          Unbalanced Power Rel. & 0.65   \\
         Spectators   & 0.54 \\
         Prejudice & 0.61   \\
           Appeal & 0.48   \\
           Sympathy & 0.71   \\
\bottomrule[1.2pt]
\end{tabularx}
\caption{Kappa IAA scores of CPCL binary and multi-class annotations.}
\label{tab:table8}

\end{table}
%IAA
%case 开始
\begin{table*}[htbp]
\centering
\subfigure{
\begin{tabular}{ 
>{\centering\arraybackslash}m{1.5cm} 
>{\centering\arraybackslash}m{6cm} 
>{\centering\arraybackslash}m{6cm} 
}
\toprule[1.2pt]
EN      & Case A(\romannumeral1) & Case B(\romannumeral1) \\ \midrule
Text & \textit {After already receiving relief funds, what else do these so-called 'poor' families think they deserve? }      & \textit {For some of these male prostitutes, the 'clients' they picked up on this corner were their only means of survival.}       \\ \midrule
Category      & "Unbalanced Power Relations","Prejudice" & "Spectator","Elicit Compassion"  \\ \midrule
Explanation   & The phrase \textcolor{red}{"so-called 'poor' families"} suggests a condescending attitude towards impoverished households, reflecting an \textcolor{red}{ unbalanced power relationship}, where those with more resources view those with less through a \textcolor{red}{biased perspective}. The tone is \textcolor{red}{dismissive and judgmental}.       &  The phrasing of this sentence suggests a \textcolor{red}{spectator's indifferent attitude} towards male prostitutes. It implies that these men have no other choice but to engage in sex work for survival. \textcolor{red}{Spectators elicit compassion for their plight} while maintaining a superior stance. The toxicity of such descriptive statements is often \textcolor{red}{complex to detect}.      \\  \midrule
Recognition Difficulty      & Middle & High \\ \midrule
Prediction    &  M-BERT:\ding{52}, RoBERTa:\ding{52}, GPT-4.0:\ding{55}, Claude-3:\ding{52}, LLaMA-2:\ding{55}, PclGPT-EN:\ding{52}      &    M-BERT:\ding{55}, RoBERTa:\ding{55}, GPT-4.0:\ding{55}, Claude-3:\ding{52}, LLaMA-2:\ding{55}, PclGPT-EN:\ding{52}   \\  
\bottomrule[1.2pt]
\end{tabular}
}
\subfigure{
\begin{CJK}{UTF8}{gbsn}
\begin{tabular}{ 
>{\centering\arraybackslash}m{1.5cm} 
>{\centering\arraybackslash}m{6cm} 
>{\centering\arraybackslash}m{6cm} 
}
\toprule[1.2pt]
CN                            & Case A(\romannumeral2) & Case B(\romannumeral2)      \\  \midrule
\multirow{2}{*}{Text} & 单亲的小孩大概率很难相处。        &农民工挣钱不容易的，确保工资该发就发吧。       \\
                               & \textbf{Translation:} \textit {Children from single-parent families often face difficulties in getting along with others.} & \textbf{Translation:} \textit {Making a living as a migrant worker is no easy task, let's make sure they receive their rightful wages.} \\  \midrule
Category      & "Unbalanced Power Relations","Prejudice" & "Appeal","Elicit Compassion"  \\ \midrule
Explanation                    &   This statement reflects an \textcolor{red}{unbalanced power relation and prejudice} against \textcolor{red}{single-parent families}. It assumes that children from such backgrounds inherently face social difficulties, \textcolor{red}{ignoring} the complexity of individual experiences and the diverse support systems that may exist.           & This superficial \textcolor{red}{appeal for fairness} to migrant workers hides implicit bias. It simplifies their fight and focuses solely on the wage situation. Due to the \textcolor{red}{lack of offensive intent}, this condescending attitude is \textcolor{red}{difficult to detect} without deeper analysis.          \\  \midrule
Recognition Difficulty      & Middle & High \\ \midrule
Prediction                    &  RoBERTa:\ding{55}, Chinese-BERT:\ding{52}, GPT-4.0:\ding{55}, Claude-3:\ding{52}, ChatGLM-3:\ding{52}, PclGPT-CN:\ding{52}   &    RoBERTa:\ding{55}, Chinese-BERT:\ding{55}, GPT-4.0:\ding{55}, Claude-3:\ding{55}, ChatGLM-3:\ding{52}, PclGPT-CN:\ding{52}     \\  \bottomrule[1.2pt]
\end{tabular}
\end{CJK}
}
\caption{Illustration of case study. We selected typical PCL samples from the English and Chinese test sets respectively. “Category” represents the fine-grained toxicity category of PCL, “Explanation" is a manual annotation analysis, and the key information is marked in \textcolor{red}{red}. \ding{52} indicates that the model has made a correct judgment, \ding{55} indicates a wrong judgment.}
\label{tab:table10}
\end{table*}
%case结束
\section{Case Study for PclGPT}
\label{app:d}
To further illustrate the rationales of PclGPT, and to determine whether the model can effectively identify the fuzzy subcategory of PCL. We selected samples from the Chinese and English test results respectively for case testing. The results are detailed in Table \ref{tab:table10}. Regarding the English part, we selected M-BERT, RoBERTa, GPT-4.0, Claude-3, LLaMA-2-7B and PclGPT-EN for comparative analysis. For Chinese data, we choose Chinese pre-trained Chinese-BERT, ChatGLM-3-6B and PclGPT-CN for comparison.
%细粒度分类开始
\begin{figure*}[h]
\centering
\hspace*{-0.7cm}
\includegraphics[width=0.75\textwidth]{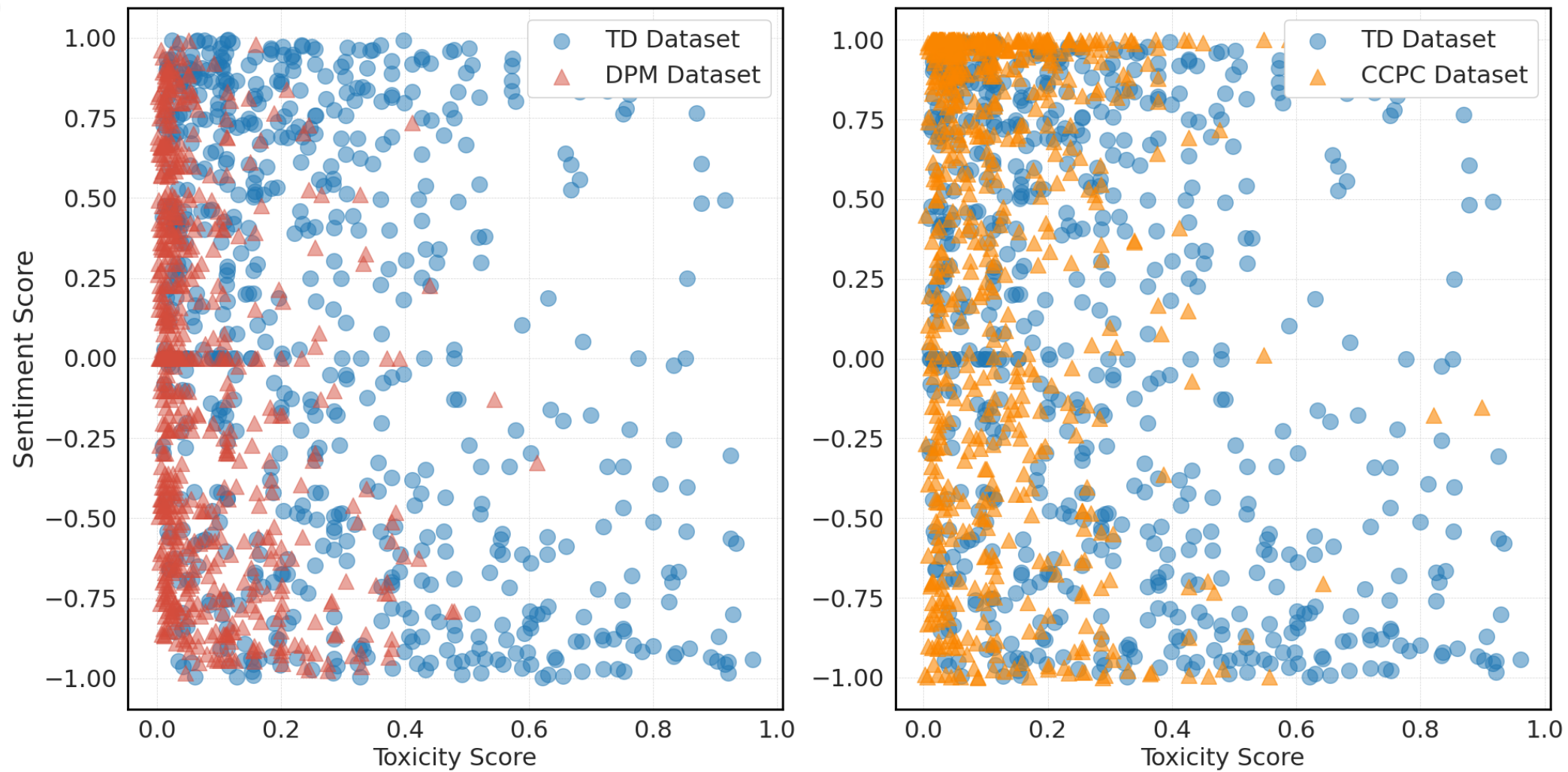}
\caption{Toxicity score scatter plots for three PCL datasets.}
\label{fig:figure6}
\end{figure*}
%细粒度分类结束
\begin{itemize}
   \item[$\bullet$]  Case A generally selects cases with "Unbalanced Power Relations" and "Prejudice" labels in PCL. In these examples, advantaged groups place themselves in a higher social status and display strong discriminatory characteristics against disadvantaged groups. For example, "so-called" in A(\romannumeral1) satirizes that poor communities should not receive subsidies, a severe expression of prejudice. A(\romannumeral2) expresses the stereotype that "children from single-parent families are difficult to get along with". The toxicity of this type of speech is apparent. Although there is no precise attack vocabulary, the models can detect it effectively. In A(\romannumeral1), most models can effectively identify the result. Similar results were obtained in A(\romannumeral2), indicating that the Chinese domain also uses the semantic information of PCL.
    
    \item[$\bullet$] The cases selected in Case B are mostly subcategories of "Spectator" and "Elicit Compassion". These categories place advantaged groups as bystanders, offering superficial opinions to solve problems or expressing sympathy for disadvantaged groups. In B(\romannumeral1), people's sympathy for the "client" is aroused through descriptive sentences, and in B(\romannumeral2), people's concern for the "migrant worker" is aroused, and people are called for guaranteed wages. The PCL toxicity of these remarks is hidden in vague expressions, and it is difficult for the model to detect the implicit toxicity. For B(\romannumeral1), only Claude-3 and PclGPT-EN correctly identified the result, while for B(\romannumeral2), only ChatGLM-3 and PclGPT-CN correctly identified the result. This demonstrates the importance of PclGPT for implicit toxicity detection.
\end{itemize}

\section{Add Implicit Interference Samples}
\label{app:e}
We conducted additional experiments to assess PclGPT's detection capabilities for implicit toxicity. As a subjective sentiment, the ambiguous part of PCL's semantic information often results in interference samples during annotation. These samples have more marginal condescending attributes, hindering the model's ability to distinguish positive samples effectively. We experimented with three dataset scenarios: without any interference samples, with a limited number of interference samples, and with all interference samples included.
%模糊情感测试图
\begin{table}[t]
\centering
\begin{tabular}{cccc}
\hline
   \textbf{Model} & \textit{S-None} &\textit{S-Few} &\textit{S-All}  \\
\hline
        
        BERT & 67.1 (0) & 67.2 (+0.1) &67.1 (-0.6)   \\
        ChatGLM  &48.1 (0)& 48.8 (+0.7) & 48.3 (-0.5) \\
        ChatGPT  & 64.3 (0) & 61.3 (-3.0)&52.4 (-8.9)\\
        GPT-4.0 & 65.5 (0) &63.2 (-2.3)&54.5 (-8.7)   \\
        PclGPT  & \textbf{67.7 (0)}& \textbf{71.5 (+3.8)}& \textbf{72.8 (+1.3)} \\
\hline
\end{tabular}
\caption{The test results of each model after gradually adding fuzzy samples. The percentage in parentheses indicates the change after addition compared with before addition.}
\label{tab:table11}
\end{table}
\textbf{ Result.} Identifying interference samples encompassing complex and implicit emotions is a difficult objective in toxicity analysis. Table \ref{tab:table11} displays the following test results. It is evident that when the number of interference intermediate examples increases, both the BERT model and the GPT baseline model experience a decrease in performance. Notably, ChatGPT and GPT-4 decline over 8\%, suggesting that they inadequately capture the condescending traits of these fuzzy cases. PclGPT is the only model that can effectively detect these interference samples in the S-Few and S-All datasets, which fully demonstrates the robust testing capabilities of our model.
\section{Toxicity Scores and Implicit Features}
\label{app:f}
Figure \ref{fig:figure6} uses a scatter plot to show the toxicity scores of the PCL test sets. The TD dataset has a smooth distribution across the entire range, while the DPM and CCPC datasets have lower average toxicity scores, with samples concentrated in low or zero-score regions. This correlates with the weaker F1 scores in the DPM and CCPC data, indicating that lower toxicity scores often align with higher implicit features, suggesting more exploration is needed for implicit toxicity. The scatter plot also shows that sentiment scores (vertical axis) have a limited impact on PCL detection, as the sentiment scores do not exhibit distinct distribution patterns.

\section{Discussion on RLHF technology}
\label{app:g}
In the early stages of the experiment, we established a Pcl-RLHF feedback dataset after the PT stage to achieve a more accurate understanding and description of toxic content. However, due to the unclear boundaries of toxicity in PCL, the model erroneously reinforced certain toxic statements during feedback ranking, leading to an increase in toxicity scores after the experiment (the average score rose from 0.37 to 0.41). Moreover, since our experiment focused more on evaluating the existing PCL classification rather than generating output, RLHF may have impacted the model’s original judgment. Therefore, RLHF was ultimately not used in this study.
%prompt1

%问卷图
\begin{figure*}[h]
\centering
\includegraphics[width=0.9\textwidth]{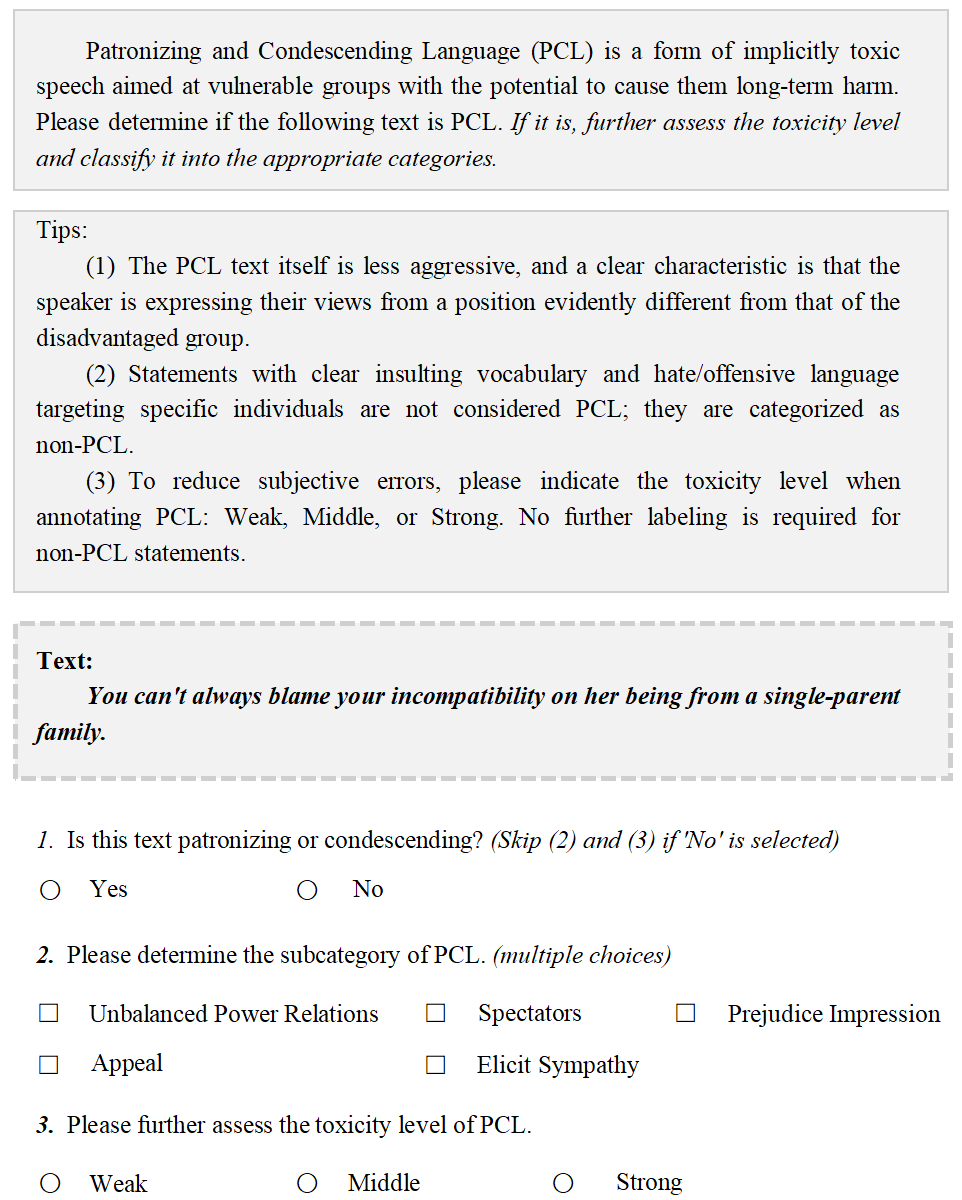} % 调整宽度
\caption{We used a web-based layered annotation questionnaire, which includes the definitions of annotations, annotation tips, and input texts. Every time we changed the text, we performed batch annotation.}
\label{fig:figure8}
\end{figure*}
%问卷图结束
\end{document}